\documentclass[sigconf]{acmart}
\usepackage{graphicx}
\usepackage{amsmath}
\usepackage{booktabs}
\usepackage[ruled]{algorithm2e}
\usepackage{multirow}
\usepackage{amsfonts}
 
\usepackage{amssymb}

\AtBeginDocument{%
  }

\setcopyright{acmlicensed}
\copyrightyear{2018}
\acmYear{2018}
\acmDOI{XXXXXXX.XXXXXXX}

\acmConference[Conference acronym 'XX]{Make sure to enter the correct
  conference title from your rights confirmation emai}{June 03--05,
  2018}{Woodstock, NY}
\acmISBN{978-1-4503-XXXX-X/18/06}

\acmSubmissionID{2024-313}

\renewcommand\footnotetextcopyrightpermission[1]{}

\begin{document}

\title{Adaptive Intra-Class Variation Contrastive Learning for Unsupervised Person Re-Identification}

\author{Lingzhi Liu}
\email{llz2021140762@bupt.edu.cn}
\affiliation{%
  \institution{Beijing University of Posts and Telecommunications}
  \country{China}}
  
\author{Haiyang Zhang}
\affiliation{%
  \institution{Beijing University of Posts and Telecommunications}
  \country{China}}

\author{Chengwei Tang}
\affiliation{%
 \institution{Xiamen University}
 \country{China}}

\author{Tiantian Zhang}
\affiliation{%
  \institution{Beijing University of Posts and Telecommunications}
  \country{China}}


\renewcommand{\shortauthors}{Trovato et al.}

\begin{abstract}
  The memory dictionary-based contrastive learning method has achieved remarkable results in the field of unsupervised person Re-ID. However, The method of updating memory based on all samples does not fully utilize the hardest sample to improve the generalization ability of the model, and the method based on hardest sample mining will inevitably introduce false-positive samples that are incorrectly clustered in the early stages of the model. Clustering-based methods usually discard a significant number of outliers, leading to the loss of valuable information. In order to address the issues mentioned before, we propose an adaptive intra-class variation contrastive learning algorithm for unsupervised Re-ID, called AdaInCV. And the algorithm quantitatively evaluates the learning ability of the model for each class by considering the intra-class variations after clustering, which helps in selecting appropriate samples during the training process of the model. To be more specific, two new strategies are proposed: Adaptive Sample Mining (AdaSaM) and Adaptive Outlier Filter (AdaOF). The first one gradually creates more reliable clusters to dynamically refine the memory, while the second can identify and filter out valuable outliers as negative samples. 
\end{abstract}

\ccsdesc[500]{Computing methodologies~Visual content-based indexing and retrieval}
\ccsdesc[500]{Computing methodologies~Image representations}
\ccsdesc[500]{Computing methodologies~Object identification}
\ccsdesc[300]{Computing methodologies~Matching}
\ccsdesc[500]{Information systems~Top-k retrieval in databases}

\keywords{Unsupervised Person Re-ID, Adaptive Intra-Class Variation, Contrastive Learning, Curriculum Learning}

\received{20 February 2007}
\received[revised]{12 March 2009}
\received[accepted]{5 June 2009}

\maketitle

\section{Introduction}
\label{sec:intro}
In recent years, there has been a rapid development of intelligent surveillance equipment and an increasing demand for public safety. As a result, a large number of cameras have been deployed in public places such as airports, communities, streets, and campuses. These camera networks typically span large geographic areas with non-overlapping coverage and generate significant amounts of surveillance video daily \cite{Ming_2022_Survey_Outlook}. The continuous increase in surveillance video and image data, along with the expensive cost of manual annotation, has led to widespread interest in the unsupervised human re-identification task. Unsupervised methods are more scalable and have a higher likelihood of being deployed in the real world compared to supervised methods.

\begin{figure}
  \centering{}
  \includegraphics[width=0.36\textheight]{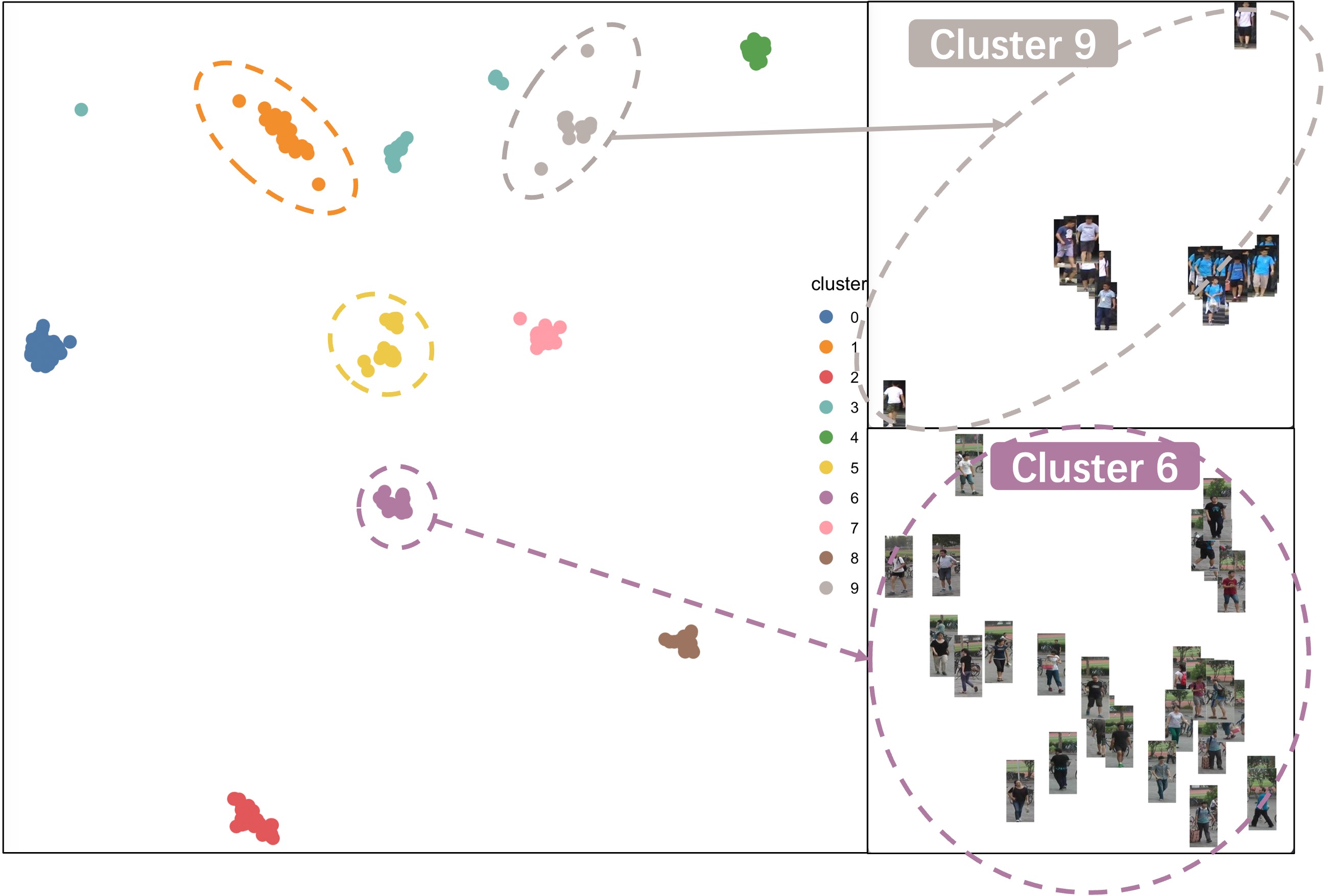}
  \caption{Using the t-SNE \cite{Van_2014_tSNE}, the results of the initial clustering of the model are depicted in the figure on the left. 10 pseudo-labels were randomly selected from the clustering results. The features were then plotted on the graph after dimensionality reduction. The figure on the right shows the two randomly selected clusters, with the scatter points replaced by the original graphs. As depicted in the graph, variations exist in the intra-class density and the distance between the most difficult samples within clusters across different clusters. Using the same learning strategy to select samples within clusters for updating features is inaccurate. Therefore, in this work, the algorithm proposed in this paper is applicable to different clusters after clustering, rather than all clusters in the entire training set.
  }
  \vspace{-0.4cm}
  \label{fig:intro}
\end{figure}

Existing unsupervised person Re-Id methods fall into the following two categories. One approach \cite{Ge_2020_MMT, Wang_2020_MMCL, Ge_2020_SpCL, Chen_2021_ABMT, Chen_2020_DCML} aims to transfer knowledge from existing labeled data to unlabeled target data, known as unsupervised domain adaptation (UDA). UDA methods typically employ a two-stage training strategy. First, the model is pre-trained on the labeled dataset from the source. Then, it is fine-tuned on the unlabeled dataset from the target in an unsupervised manner. Another type of work \cite{Dai_2021_CC, Chen_2021_ICE, Zhang_2022_ISE, Wu_2022_MCRN, Wang_2021_CAP, Cheng_2022_HDCRL} relies on unsupervised learning (USL) to learn feature representations of images from unlabeled data. They generally assign pseudo labels to completely unlabeled training samples. And then gradually use pseudo labels for classification or metric learning.

Our work mainly focuses on pure unsupervised person Re-ID. The performance of unsupervised methods relies on learning feature representations. Recently, the state-of-the-art (SOTA) feature representation learning mainly uses Memory dictionary to store all instance features. We categorize the methods of updating the memory dictionary in unsupervised Re-ID based on contrastive learning into the following two categories: (1) \textbf{Using the average features of all samples within the cluster to update}. In \cite{Ge_2020_SpCL} and \cite{Dai_2021_CC}, all pseudo-labels are regarded as clean data during optimization, without considering quality diversity. This leads to slow convergence and limited generalization ability. (2) \textbf{Using the hardest sample to update}. Certain methods, such as \cite{Chen_2021_ICE} and \cite{Cheng_2022_HDCRL}, employ hard mining during the memory database update process. This approach prioritizes the hardest samples within the class, bringing them closer to the normal samples. During the initial stages of model training, the hardest positive samples within a cluster may be noise samples. This is because the ground truth is not available, which can contaminate their corresponding cluster feature representations  and have a negative impact on the model's performance. This problem is particularly serious in large-scale Re-ID datasets, such as MSMT17 \cite{Wei_2018_MSMT17}.

In order to address the problems that exist in the above two kinds of methods and select the most suitable samples to update the memory based on the current capabilities of the model, we utilize the concept of curriculum learning \cite{Bengio_2009_Learning}. This involves gradually increasing the training difficulty of the model by introducing more complex samples over time. The key to effective curriculum learning is determining the appropriate order of samples. This ensures that the model experiences a smooth transition during the learning process, avoiding both premature difficulty and premature saturation. This process usually requires dynamic adjustments based on model performance and training progress to ensure optimal training results.

In traditional curriculum learning \cite{Spitkovsky_2010_Baby_Steps}, a linear scheduling strategy is generally used to start sampling from the simplest data and gradually increase the difficulty until the most difficult. In self-paced learning \cite{Kumar_2010_Self}, the model is trained at each iteration using the proportion of data with the lowest training loss. Both learning methods work on the entire training dataset. However, in the field of unsupervised Re-ID, as shown in \ref{fig:intro} , there are variations in intra-class density and distance between the hardest positive samples across different clusters. It would be inaccurate to use the same learning strategy to select samples within a cluster for updating clustering features. Therefore, in this work, the algorithm we propose works on different categories after clustering, rather than the entire training set.

In terms of determining the model's learning ability for different clusters, we drew inspiration from \cite{Zhou_2023_AdaSP} and evaluated the variations within each cluster by measuring the similarity between the hardest positive pair and the least-hardest pair. This allows us to infer the current performance of the model. If there is a significant variation within the class, it indicates that the model's ability is weak. Conversely, if the intra-class variation is small, it suggests that the model's ability is adequate. 
When the model's ability is sufficient, we utilize the hardest sample to update the memory. 
When the model's capability is insufficient, select an appropriate sample for updating based on its performance. 

In addition, a large number of outliers may be generated during the clustering of image features. In order to simplify the processing and ensure the reliability of the generated pseudo labels, existing methods \cite{Chen_2021_ICE, Dai_2021_CC, Zhang_2022_ISE, Wang_2021_CAP} simply discard outliers that contain contain a wealth of valuable information. Although outliers always contain many challenging training samples, they are valuable for person Re-ID tasks. 
Simply discarding such training samples can severely impair model performance. However, outliers cannot be indiscriminately stored in the memory dictionary either. In the early stages of the model, most outliers are likely due to inadequate feature extraction capability, resulting in inaccurate clustering. Therefore, we have also proposed a strategy that dynamically incorporates outliers into the memory dictionary based on the current capability of the model.

In summary, our proposed Cluster Contrast for unsupervised Re-ID has the following contributions:
\begin{itemize}
\item

We propose an algorithm \textbf{AdaInCV} that utilizes the intra-class variations after clustering to assess the learning capability of the model for each class \textbf{separately}. This allows for the selection of appropriate samples during the training process of the model.
\item
More specifically, we propose two new strategies: \textbf{AdaSaM} enables the model to select samples of appropriate difficulty based on the learning ability of each cluster in order to update the memory; \textbf{AdaOF} utilizes the learning ability of the model across the entire dataset to select appropriate outliers as negative samples, thereby enhancing contrastive learning.
\item
Extensive experiments on two popular large-scale Re-ID benchmarks demonstrate that our algorithm \textbf{AdaInCV} outperforms previous state-of-the-art methods and significantly improves the performance of unsupervised person Re-ID.
\end{itemize}

\section{Related Work}
\label{sec:related work}

\subsection{Deep Unsupervised Person Re-ID}
In recent years, unsupervised person re-id methods can be broadly categorized into unsupervised domain adaptation (UDA) and fully unsupervised methods. In UDA-based methods, several studies utilize transformations of image semantic information to minimize the domain gap between the source domain and the target domain. This is achieved by converting source domain images into the style of target domain images \cite{Chen_2021_ICE}, while still retaining the identity information of the individuals in the source domain. Alternatively, some methods employ domain-specific networks to utilize complementary information between domains, enhancing the model's performance through domain generation methods.

Fully unsupervised person re-id methods train models on unlabeled datasets. State-of-the-art unsupervised learning pipelines typically involve three stages: pseudo-label generation, memory dictionary initialization, and neural network training \cite{Dai_2021_CC}. Previous work has made various improvements at different stages of this pipeline. 
The SpCL \cite{Ge_2020_SpCL} introduces a self-paced contrastive learning framework with hybrid memory. The proposed self-paced method gradually creates more reliable clusters to refine the hybrid memory and learning targets.
ICE \cite{Chen_2021_ICE} uses pairwise similarity ranking to mine the hardest samples for hard instance contrast, thereby minimizing intra-class variance, and pairwise similarity scores as soft pseudo-labels to enhance the consistency between the data augmented and the original instances.
HHCL \cite{Hu_2021_HHCL} combines cluster-level contrastive loss with instance-level contrastive loss and updates the corresponding cluster representation with the features of the hardest positive instance in each sample. This approach aims to further mine the discriminant information.
ISE \cite{Zhang_2022_ISE} reduces the impact of noisy clusters on the model by employing a Progressive Linear Interpolation (PLI) strategy to generate support samples from actual samples and their neighboring clusters in the embedding space. 

Despite significant performance improvements achieved by enhancing various stages of the pipeline, these methods often overlook a crucial aspect, which limits their further improvement. Specifically, a single filtering mode is always used when updating the features in the memory bank, while ignoring the intra-class variations between different clusters. Appropriate samples should be selected for updating based on the model's learning ability for different clusters.

\subsection{Memory dictionary}
In the current field of unsupervised representation learning for computer vision, contrastive learning achieves state-of-the-art results. 
MoCo \cite{He_2020_MoCo} introduced a dynamic dictionary with a queue and a momentum-updated encoder, aiming to build a large and consistent dictionary for better contrastive learning.
In SimCLR \cite{Chen_2020_SimCLR}, achieving good results has also been possible by directly using large batch size for dictionary lookup tasks.
BYOL \cite{Grill_2020_BYOL} directly removes the contrastive learning of the negative sample part based on MoCo. It adds the Predictor module after the positive sample projection to predict image features. This is achieved through the asymmetric tissue model of the upper and lower branching structure, resulting in better results compared to previous work.

However, the above method treats each unlabeled sample as a distinct class to learn instance discriminative representations. This approach is not suitable for a fine-grained person ReID task.

Therefore, SpCL \cite{Ge_2020_SpCL} redefines the contrastive learning method in the Re-ID task by calculating the loss at the cluster level and updating the memory dictionary at the instance level. Building on this, CC \cite{Dai_2021_CC} proposes to calculate the loss at the cluster level and update the memory dictionary. So far, many methods based on the aforementioned two works have achieved impressive results.

\subsection{Curriculum Learning}
Curriculum learning (CL) is a training strategy that trains a machine learning model from easier data to harder data, which imitates the meaningful learning order in human curricula \cite{Wang_2021_Curriculum}. The original concept of CL is first proposed by \cite{Bengio_2009_Learning}. The advantages of applying CL training strategies to miscellaneous real-world scenarios can be mainly summarized as improving the model performance on target tasks and accelerating the training process, which cover the two most significant requirements in major machine learning research. In this paper, the concept of Curriculum learning is introduced for the first time into the task of person Re-ID. This approach accelerates the convergence speed of the model and ultimately achieves the state-of-the-art (SOTA) performance.

\section{Methodology}

\begin{figure*}[t]
  \centering{}
  \includegraphics[width=0.73\textheight]{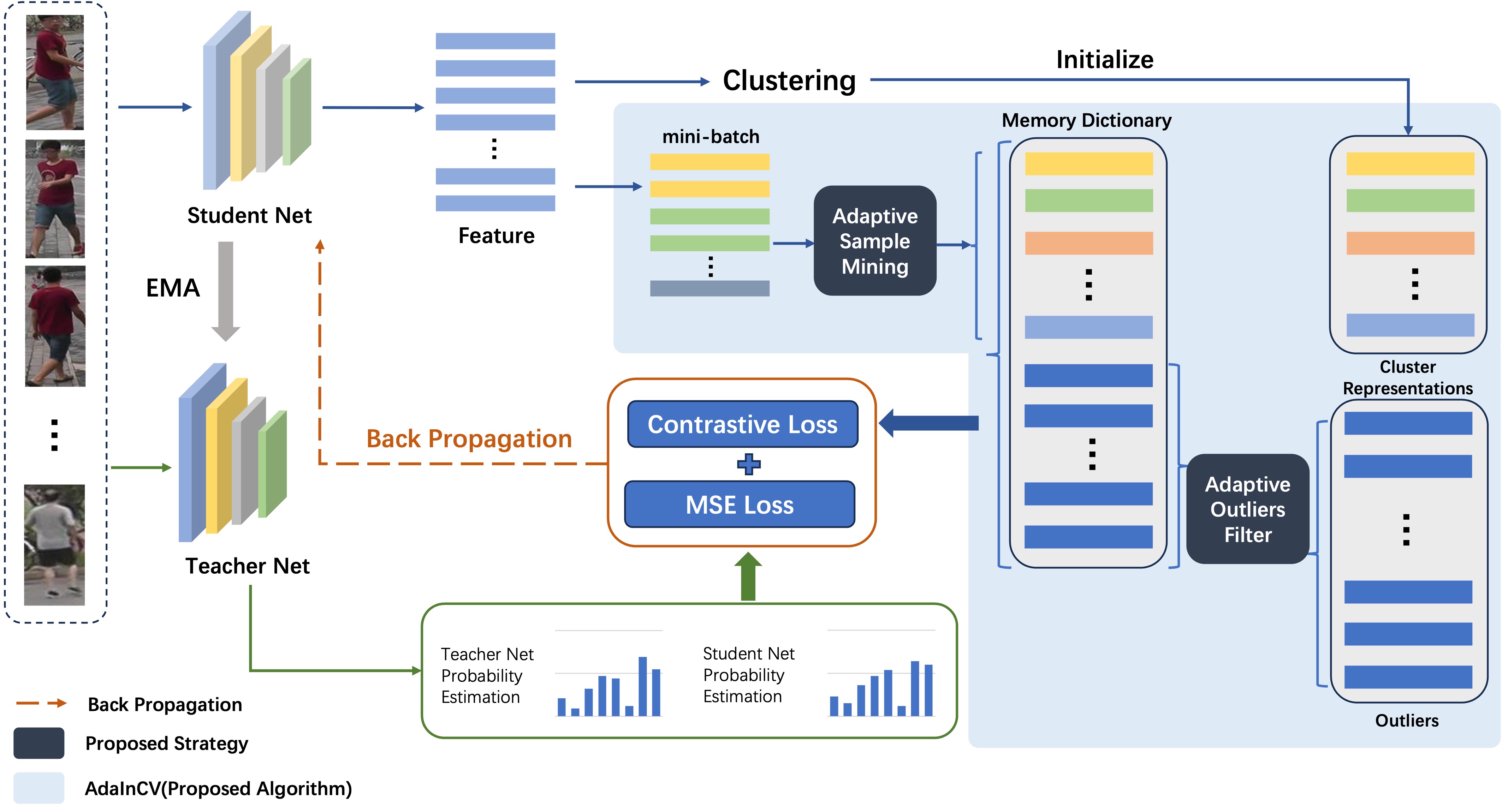}
  \caption{Illustration of the framework with Adaptive Sample Mining and Adaptive Outliers Filter, which work together to provide reliable positive and negative samples for contrastive learning. We propose a training process for adaptive sample mining, which utilizes the idea of curriculum learning to enable the model to select samples of the suitable difficulty level for updating the corresponding clustering features based on the learning capacity of each cluster.}
  \label{fig:model}
\end{figure*}

In this section, we will mathematically model the entire problem and first introduce our overall approach in \ref{sec:Overview}, explaining how to incorporate the concept of contrastive learning into our work. Then, in \ref{sec:adaModelDiff}, we will provide a detailed explanation of how to evaluate the strength of the model's ability and derive the calculation formula for adaptive similarity from a mathematical perspective. After that, the details of momentum update and its working principle are explained in \ref{sec:adaSampleMine}. Finally, the outlier is adaptively selected as negative samples in \ref{sec:adaOutlierFilt} to enhance the effect of contrast learning. An overview of the pipeline is shown in \ref{fig:model}.

\subsection{Overview}
\label{sec:Overview}
Given an unlabeled data set $\mathcal{X} = \{x_1, x_2, ...., x_N\}$, where $x_i$ represents the $i$-th image sample and $N$ represents the number of samples in the dataset. Our goal is to train a robust model $f_{\theta}(\cdot)$ on unlabeled $\mathcal{X}$ in order to minimize the variations between the features of the same person captured from various viewpoints or cameras with different settings. Additionally, we aim to ensure that the features of different person are as far apart as possible.

For each sample in the dataset, the Re-ID model $f_{\theta}(\cdot)$ will generate a corresponding feature embedding. Although the initial model is not optimized for the Re-ID task, it still has the most basic level of discriminability. The DBSCAN clustering algorithm is then applied to cluster similar features together and assign pseudo labels to them. The cluster feature representation is initialized and calculated as the average feature vector of the samples in each cluster. The Memory dictionary is then initialized using these cluster features and their corresponding pseudo labels. The formula of $k$-th cluster features $c_k$ is expressed as follows:
\begin{equation}
c_k = \frac{1}{|\mathcal{H}_k|}\sum_{f_\theta(x_i) \in \mathcal{H}_k} f_\theta(x_i)
\label{cluster_features}
\end{equation}
where $\mathcal{H}_k$ denotes the $k$-th cluster set that contains all the feature embedding within cluster $k$ and $|\cdot|$ denotes the number of features in the set

Due to the inherent characteristics of the DBScan algorithm, a significant number of outliers will be generated during the clustering process. We will filter the outliers and select suitable outliers to add to the Memory dictionary, considering the current capabilities of the model, as negative samples. During the training phase, we will compute the HybridNCE loss between the query image features and all cluster representations in the dictionary in order to train the network. At the same time, the appropriate query sample feature is selected to update the corresponding dictionary feature based on the intra-class variations of the class to which the query feature belongs. HybridNCE Loss was proposed mainly to increase the number of negative samples, ensuring a more stable and balanced learning process, $\mathcal{L}_{hybrid}$ is formulated as follows:
\begin{equation}
 -log\frac{exp(\langle q \cdot \Phi_k^+ \rangle/\tau)}{\sum_{k=1}^{n_c}exp(\langle q \cdot \Phi_k \rangle / \tau) + \sum_{k=1}^{n_o}exp(\langle q \cdot \mathcal{I}_k \rangle / \tau)}
\label{HybridNCE}
\end{equation}
where $\Phi_k$ is the unique representation vector of the $k$-th cluster. $\mathcal{I}_k$ is the outliers filtered based on the current model's ability. For any query person image $q$, $\Phi_k^+$ represents the positive cluster feature to which $q$ belongs. The temperature $\tau$ is empirically set to 0.05, and $\langle \cdot, \cdot \rangle$ denotes the inner product between two feature vectors, used to measure their similarity. $n_c$ is the number of clusters and $n_o$ is the number of un-clustered instance.

In addition, inspired by MoCo \cite{He_2020_MoCo}, we design our proposed method with a teacher network $f_{\theta_t}(\cdot)$ and a student network $f_{\theta_s}(\cdot)$. The student network is updated by back-propagation and the teacher has the same structure as the student network, but is exponential moving average (EMA) updated by the student model’s weights. we learn to match these class probability distributions P by minimizing the Mean Square Loss $\mathcal{L}_{mse}$ in the self-supervised manner.
\begin{equation}
 \mathcal{L}_{mse} = \Vert P^s(x_i) - P^t(x_i) \Vert
\label{Probability Distillation}
\end{equation}
where $P_s$ and $P_t$ are the probability distributions generated by student and teacher network respectively. 

Our optimization goal is to minimize the overall loss, which can be formulated as:
\begin{equation}
 \mathcal{L}_{total} = \mathcal{L}_{hybrid} + \lambda_m\mathcal{L}_{mse}
\label{Loss}
\end{equation}
where $\lambda_m$ is the hyper-parameter to balance these two objectives.

\subsection{Adaptive Model Capability Acquisition}
\label{sec:adaModelDiff}

Although the Average strategy used to update the Memory dictionary in SpCL and CC has achieved impressive results, it ignores the need to mine more challenging samples in the later stages of the model. On the other hand, in HHCL \cite{Hu_2021_HHCL}, ICE, and HDCRL, the hardest positive sample in the class is utilized to update the memory dictionary, resulting in significant improvements. However, this approach does not take into account the relatively weak capabilities of the model in the early stages, which can introduce wrong samples during the training process. To address these issues, we have adopted the concept of contrastive learning, gradually increasing the difficulty of training the model by introducing the complexity of the sample over time.

There are N identities in a mini-batch, where each identity has K positive samples. We represent the result of normalizing the characteristics of the $k$-th instance in class $i$ as $\mathcal{F}_i^k = \frac{f_i^k}{\Vert f_i^k \Vert}$.

Inspired by geometric insight, we find that the hardest pair of positive samples in the class determines the radius of the largest hypersphere covering all samples within the class. The similarity of the most difficult positive sample pairs in class $i$ can be obtained using the following formula:
\begin{equation}
 Sim_{h, i}^+ \approx -\tau log(\sum_{n=1}^K\sum_{m=1}^K e^{-\frac{\mathcal{F}_i^n \mathcal{F}_i^m}{\tau}})
\label{hardest pair}
\end{equation}

Each positive instance, when paired with its furthest positive sample, can form an intra-class hypersphere with varying radii. This is similar to a hypersphere formed using the radius of the hardest positive sample. The sphere formed by a sample can encompass all positive samples. The similarity of the $n$-th positive pair can be determined using the following equation:
\begin{equation}
 Sim_{n, i}^+ \approx -\tau log(\sum_{m=1}^K e^{-\frac{\mathcal{F}_i^n \mathcal{F}_i^m}{\tau}})
\label{nth positive pair}
\end{equation}

It is extremely challenging to correct the visual discrepancies between pairs of samples that are caused by complex lighting, dense occlusion, or different viewpoints. Especially when the model's capability is insufficient in the early stages of training, the hardest positive sample pairs are likely to be incorrectly clustered samples. This can lead to conflicts with the corresponding features in the memory dictionary, ultimately misleading the process of feature learning. Therefore, the least hardest positive sample pair should be considered. The similarity of the least difficult positive sample pair is calculated as follows:
\begin{equation}
\begin{split}
 Sim_{lh, i}^+ \approx \tau log(\sum_{n=1}^K e^{\frac{Sim_{n, i}^+}{\tau}}) \\
            = \tau log(\sum_{n=1}^K\frac{1}{\sum_{m=1}^K e^{-\frac{\mathcal{F}_i^n \mathcal{F}_i^m}{\tau}}})
\end{split}
\label{least hardest positive pair}
\end{equation}

For categories with large intra-class changes, we believe that the current model's ability is relatively weak and can be addressed by mining the least hardest positive samples and using simpler samples. For data with relatively small intra-class changes, our task model's ability is sufficient, but the training process can be accelerated by incorporating the hardest positive samples and harder samples. Unfortunately, the fixed sampling strategy cannot dynamically handle these two cases. We assume that the variation between the largest and smallest intra-class hyperspheres can be used to approximate the feature variation within a class. Based on this, we propose an adaptive weighting strategy that dynamically balances the two positive pairs according to different intra-class variations.

Therefore, to weigh the variations in the training phase, we compute the harmonic average of the hardest and least hardest similarities of class $i$ in the mini-batch:
\begin{equation}
 h_i = \frac{2Sim_{lh, i}^+ Sim_{h, i}^+}{Sim_{lh, i}^+ + Sim_{h, i}^+}
\label{harmonic average}
\end{equation}

Because the harmonic mean property satisfies our proposed requirements, we adopt the harmonic mean of similarities as an adaptive weight to balance the hardest and least hardest pairs of each class:
\begin{equation}
 \alpha_i = \begin{cases}
    h_i, & Sim_{h, i}^+ \geq 0\\
    0, & Sim_{h, i}^+ < 0
    \end{cases}
\label{alpha weight}
\end{equation}

The weighted positive similarity can be given by:
\begin{equation}
 Sim_i^+ = \alpha_i Sim_{h, i}^+ + (1 - \alpha_i) Sim_{lh, i}^+
\label{weighted positive pair}
\end{equation}

\subsection{Adaptive Sample Mining}
\label{sec:adaSampleMine}

Different from the instance-level memory dictionary, we mainly store the features of each cluster in the memory-based dictionary, and calculate the variation between samples within the class corresponding to the pseudo-label according to the weighted positive sample pair similarity obtained in \ref{sec:adaModelDiff}. 
\begin{equation}
diff_i = \frac{Sim_i^+ - Sim_{h, i}^+}{Sim_{lh, i}^+ - Sim_{h, i}^+}
\label{difference}
\end{equation}

If the intra-class variation is small, use the hardest sample to update the corresponding Memory dictionary. If the intra-class variation is large, calculate the sample that best fits the current difficulty level according to diff to update. The specific formula is as follows:
\begin{equation}
 \beta_i = \begin{cases}
    1, & round(\frac{Sim_{lh, i}^+}{Sim_{h, i}^+}) == 1\\
    diff_i, & Others
    \end{cases}
\label{Memory update}
\end{equation}

We rewrite the momentum updating formula as follow:
\begin{equation}
\Phi_k = m \Phi_k + (1 - m) q_{rank_i}^{\beta * N}
\label{momentum update}
\end{equation}

where $q_{rank_i}^{\beta * N}$ represents the $(\beta * N)$-th sample feature obtained by sorting from largest to smallest after calculating the similarity between the $i$-th cluster feature and the sample.

\subsection{Adaptive Outlier Filter}
\label{sec:adaOutlierFilt}

The average of the diff values in all clusters $diff_{global}$ is used as an indicator to measure the difficulty of the current model. Based on this indicator, outliers are reintegrated into the Memory dictionary to increase the number of negative samples and enhance contrastive learning. This is achieved through computational sampling. The distance between samples and (clustering features + outliers) is sorted from largest to smallest. In the early stages of the model, samples with further distances should be selected as much as possible, as the model's capability is not yet sufficient. At this time, the samples that are further away are more likely to be negative samples. As the model's ability strengthens, all outliers will be absorbed from far to near.
\begin{equation}
diff_{global} = \frac{\sum_{i=1}^N diff_i}{N}
\label{difference mean}
\end{equation}

\begin{table*}[t]
\resizebox{\linewidth}{!}{
\begin{tabular}{lccccccccccc}
\hline
\multicolumn{1}{c|}{\multirow{2}{*}{Method}} & \multicolumn{1}{c|}{\multirow{2}{*}{Reference}} & \multicolumn{5}{c|}{Market-1501}                                                     & \multicolumn{5}{c}{MSMT17}                                          \\ \cline{3-12} 
\multicolumn{1}{c|}{}                        & \multicolumn{1}{c|}{}                           & \multicolumn{1}{c|}{Source} & mAP    & Top1     & Top5     & \multicolumn{1}{c|}{Top10}    & \multicolumn{1}{c|}{Source}     & mAP    & Top1     & Top5     & Top10    \\ \hline 
\multicolumn{10}{l}{Unsupervised Domain Adaptation }   \\  
\cline{1-10} 
\hline
\noalign{\smallskip}

\multicolumn{1}{l|}{AD-Cluster \cite{Zhai_2020_Ad-cluster}}         & \multicolumn{1}{c|}{CVPR'20}                    & \multicolumn{1}{c|}{Duke}   & 68.3   & 86.7   & 94.4   & \multicolumn{1}{c|}{96.5}   & \multicolumn{1}{c|}{-}          & -      & -      & -      & -      \\
\multicolumn{1}{l|}{MMT \cite{Ge_2020_MMT}}                         & \multicolumn{1}{c|}{ICLR'20}                    & \multicolumn{1}{c|}{Duke}   & 71.2   & 87.7   & 94.9   & \multicolumn{1}{c|}{96.9}   & \multicolumn{1}{c|}{Market} & 22.9   & 49.2   & 63.1   & 68.8   \\
\multicolumn{1}{l|}{DG-Net++ \cite{Zou_2020_DG-Net++}}              & \multicolumn{1}{c|}{ECCV'20}                    & \multicolumn{1}{c|}{Duke}   & 61.7   & 82.1   & 90.2   & \multicolumn{1}{c|}{92.7}   & \multicolumn{1}{c|}{Market} & 22.1   & 48.4   & 60.9   & 66.1   \\
\multicolumn{1}{l|}{SpCL \cite{Ge_2020_SpCL}}                       & \multicolumn{1}{c|}{NeurIPS'20}                 & \multicolumn{1}{c|}{Duke}   & 76.7   & 90.3   & 96.2   & \multicolumn{1}{c|}{97.7}   & \multicolumn{1}{c|}{Market} & 26.8   & 53.7   & 65.0   & 69.8   \\
\multicolumn{1}{l|}{UNRN \cite{Zheng_2021_UNRN}}                    & \multicolumn{1}{c|}{AAAI'21}                    & \multicolumn{1}{c|}{Duke}   & 78.1   & 91.9   & 96.1   & \multicolumn{1}{c|}{97.8}   & \multicolumn{1}{c|}{Market} & 25.3   & 52.4   & 64.7   & 69.8   \\
\multicolumn{1}{l|}{HGA \cite{Zhang_2021_HGA}}                      & \multicolumn{1}{c|}{AAAI'21}                    & \multicolumn{1}{c|}{Duke}   & 70.3   & 89.5   & 93.6   & \multicolumn{1}{c|}{95.5}   & \multicolumn{1}{c|}{Market} & 25.5   & 55.1   & 61.2   & 65.5   \\
\multicolumn{1}{l|}{GLT \cite{Zheng_2021_GLT}}                      & \multicolumn{1}{c|}{CVPR'21}                    & \multicolumn{1}{c|}{Duke}   & 79.5   & 92.2   & 96.5   & \multicolumn{1}{c|}{97.8}   & \multicolumn{1}{c|}{Market} & 26.5   & 56.6   & 67.5   & 72.0   \\
\multicolumn{1}{l|}{GCL \cite{Chen_2021_GCL}}                       & \multicolumn{1}{c|}{CVPR'21}                    & \multicolumn{1}{c|}{Duke}   & 75.4   & 90.5   & 96.2   & \multicolumn{1}{c|}{97.8}   & \multicolumn{1}{c|}{Market} & 27.0   & 51.1   & 63.9   & 69.9   \\
\multicolumn{1}{l|}{RDSBN+MDIF \cite{Bai_2021_RDSBN}}               & \multicolumn{1}{c|}{CVPR'21}                    & \multicolumn{1}{c|}{Duke}   & 81.5   & 92.9   & 97.6   & \multicolumn{1}{c|}{98.4}   & \multicolumn{1}{c|}{Market} & 30.9   & 61.2   & 73.1   & 77.4   \\
\multicolumn{1}{l|}{MCRN \cite{Wu_2022_MCRN}}                       & \multicolumn{1}{c|}{AAAI'22}                    & \multicolumn{1}{c|}{Duke}   & 83.8   & 93.8   & 97.5   & \multicolumn{1}{c|}{98.5}   & \multicolumn{1}{c|}{Market} & 32.8   & 64.4   & 75.1   & 79.2   \\
\multicolumn{1}{l|}{CaCL \cite{Lee_2023_CaCL}}                       & \multicolumn{1}{c|}{ICCV'23}                    & \multicolumn{1}{c|}{MSMT17}   & 84.7   & 93.8   & 97.7   & \multicolumn{1}{c|}{98.6}   & \multicolumn{1}{c|}{Market} & 36.5   & 66.6   & 75.3   & 80.1   \\
\multicolumn{1}{l|}{DUCL+PAFR \cite{Liu_2023_DUCL}}                       & \multicolumn{1}{c|}{ICASSP'23}                    & \multicolumn{1}{c|}{Duke}   & 84.0   & 93.9   & 97.5   & \multicolumn{1}{c|}{98.6}   & \multicolumn{1}{c|}{Market} & 33.9   & 65.0   & 74.8   & 79.7   \\

\noalign{\smallskip}
\hline
\multicolumn{10}{l}{Fully Unsupervised (camera-aware)} \\ \cline{1-10} 
\hline
\noalign{\smallskip}
\multicolumn{1}{l|}{ICE† \cite{Chen_2021_ICE}}                      & \multicolumn{1}{c|}{ICCV'21}                    & \multicolumn{1}{c|}{None}   & 82.3   & 93.8   & 97.6   & \multicolumn{1}{c|}{98.4}   & \multicolumn{1}{c|}{None}       & 38.9   & 70.2   & 80.5   & 84.4   \\
\multicolumn{1}{l|}{CAP† \cite{Wang_2021_CAP}}                      & \multicolumn{1}{c|}{AAAI'21}                    & \multicolumn{1}{c|}{None}   & 79.2   & 91.4   & 96.3   & \multicolumn{1}{c|}{97.9}   & \multicolumn{1}{c|}{None}       & 36.9   & 67.4   & 78.0   & 81.4   \\
\multicolumn{1}{l|}{PPLR† \cite{Cho_2022_PPLR}}                      & \multicolumn{1}{c|}{CVPR’22}                    & \multicolumn{1}{c|}{None}   & 84.4   & 94.3   & 97.8   & \multicolumn{1}{c|}{98.6}   & \multicolumn{1}{c|}{None}       & 42.2   & 73.3   & 83.5   & 86.5   \\

\noalign{\smallskip}
\hline
\multicolumn{10}{l}{Fully Unsupervised (camera-agnostic)} \\ \cline{1-10} 
\hline
\noalign{\smallskip}

\multicolumn{1}{l|}{BUC \cite{Lin_2019_BUC}}                        & \multicolumn{1}{c|}{AAAI'19}                    & \multicolumn{1}{c|}{None}   & 29.6   & 61.9   & 73.5   & \multicolumn{1}{c|}{78.2}   & \multicolumn{1}{c|}{None}       & -      & -      & -      & -      \\
\multicolumn{1}{l|}{SSL \cite{Lin_2020_SSL}}                        & \multicolumn{1}{c|}{CVRP'20}                    & \multicolumn{1}{c|}{None}   & 37.8   & 71.7   & 83.8   & \multicolumn{1}{c|}{87.4}   & \multicolumn{1}{c|}{None}       & -      & -      & -      & -      \\
\multicolumn{1}{l|}{JVTC \cite{Li_2020_JVTC}}                       & \multicolumn{1}{c|}{ECCV'20}                    & \multicolumn{1}{c|}{None}   & 41.8   & 72.9   & 84.2   & \multicolumn{1}{c|}{88.7}   & \multicolumn{1}{c|}{None}       & 15.1   & 39.0   & 50.9   & 56.8   \\
\multicolumn{1}{l|}{MMCL \cite{Wang_2020_MMCL}}                     & \multicolumn{1}{c|}{CVPR'20}                    & \multicolumn{1}{c|}{None}   & 45.5   & 80.3   & 89.4   & \multicolumn{1}{c|}{92.3}   & \multicolumn{1}{c|}{None}       & 11.2   & 35.4   & 44.8   & 49.8   \\
\multicolumn{1}{l|}{HCT \cite{Zeng_2020_HCT}}                       & \multicolumn{1}{c|}{CVPR'20}                    & \multicolumn{1}{c|}{None}   & 56.4   & 80.0   & 91.6   & \multicolumn{1}{c|}{95.2}   & \multicolumn{1}{c|}{None}       & -      & -      & -      & -      \\
\multicolumn{1}{l|}{CycAs \cite{Wang_2020_CycAs}}                   & \multicolumn{1}{c|}{ECCV'20}                    & \multicolumn{1}{c|}{None}   & 64.8   & 84.8   & -      & \multicolumn{1}{c|}{-}      & \multicolumn{1}{c|}{None}       & 26.7   & 50.1   & -      & -      \\
\multicolumn{1}{l|}{SpCL \cite{Ge_2020_SpCL}}                       & \multicolumn{1}{c|}{NeurIPS'20}                 & \multicolumn{1}{c|}{None}   & 73.1   & 88.1   & 95.1   & \multicolumn{1}{c|}{97.0}   & \multicolumn{1}{c|}{None}       & 19.1   & 42.3   & 55.6   & 61.2   \\
\multicolumn{1}{l|}{GCL \cite{Chen_2021_GCL}}                       & \multicolumn{1}{c|}{CVPR'21}                    & \multicolumn{1}{c|}{None}   & 66.8   & 87.3   & 93.5   & \multicolumn{1}{c|}{95.5}   & \multicolumn{1}{c|}{None}       & 21.3   & 45.7   & 58.6   & 64.5   \\
\multicolumn{1}{l|}{ICE \cite{Chen_2021_ICE}}                       & \multicolumn{1}{c|}{ICCV'21}                    & \multicolumn{1}{c|}{None}   & 79.5   & 92.0   & 97.0   & \multicolumn{1}{c|}{98.1}   & \multicolumn{1}{c|}{None}       & 29.8   & 59.0   & 71.7   & 77.0   \\
\multicolumn{1}{l|}{HHCL \cite{Hu_2021_HHCL}}                       & \multicolumn{1}{c|}{Arxiv'21}                   & \multicolumn{1}{c|}{None}   & 84.2   & 93.4   & 97.7   & \multicolumn{1}{c|}{98.5}   & \multicolumn{1}{c|}{None}       & -   & -   & -   & -   \\
\multicolumn{1}{l|}{MCRN \cite{Wu_2022_MCRN}}                       & \multicolumn{1}{c|}{AAAI'22}                    & \multicolumn{1}{c|}{None}   & 80.8   & 92.5   & -      & \multicolumn{1}{c|}{-}      & \multicolumn{1}{c|}{None}       & 31.2   & 63.6   & -        & -        \\
\multicolumn{1}{l|}{ISE \cite{Zhang_2022_ISE}}                      & \multicolumn{1}{c|}{CVPR'22}                    & \multicolumn{1}{c|}{None}   & 84.7   & 94.0   & 97.8   & \multicolumn{1}{c|}{98.8}   & \multicolumn{1}{c|}{None}       & 35.0   & 64.7   & 75.5   & 79.4   \\
\multicolumn{1}{l|}{ClusterContrast \cite{Dai_2021_CC}}             & \multicolumn{1}{c|}{ACCV'22}                    & \multicolumn{1}{c|}{None}   & 82.6   & 93.0   & 97.0   & \multicolumn{1}{c|}{98.1}   & \multicolumn{1}{c|}{None}       & 27.6   & 56.0   & 66.8   & 71.5   \\
\noalign{\smallskip}
\hline
\noalign{\smallskip}
\multicolumn{1}{l|}{HDCRL \cite{Cheng_2022_HDCRL} (Our Baseline)}    & \multicolumn{1}{c|}{TIP'22}                     & \multicolumn{1}{c|}{None}   & 84.5   & 93.5   & 97.6   & \multicolumn{1}{c|}{98.6}   & \multicolumn{1}{c|}{None}       & 20.7   & 43.8   & 55.1   & 60.1   \\
\multicolumn{1}{l|}{\textbf{Ours}}                                 & \multicolumn{1}{c|}{This paper}                 & \multicolumn{1}{c|}{None}   & 86.1 & 94.2 & 97.8 &  \multicolumn{1}{c|}{98.7} & \multicolumn{1}{c|}{None}       & 36.0 & 65.6 & 76.2 & 79.9 \\

\multicolumn{1}{l|}{\textbf{Ours + GeM}}                           & \multicolumn{1}{c|}{This paper}                 & \multicolumn{1}{c|}{None}   & \textbf{87.4} & \textbf{94.9} & \textbf{98.1} &  \multicolumn{1}{c|}{\textbf{98.8}} & \multicolumn{1}{c|}{None}       & \textbf{38.8} & \textbf{69.8} & \textbf{79.6} & \textbf{83.0} \\

\noalign{\smallskip}
\hline
\noalign{\smallskip}

\end{tabular}
}
\caption{Comparison of Re-ID methods on Market-1501 and MSMT17 datasets. The best USL results \textbf{without camera} information are marked with \textbf{bold}. † indicates using the additional camera knowledge (camera-aware), while the rest is camera-agnostic.}
\label{tab:sota}
\end{table*}

\section{Experiments}
\label{sec:experiments}
\subsection{Datasets and Evaluation Protocol}
\textbf{Datasets.}
We evaluate our proposed method on two large-scale person re-identification (Re-ID) benchmarks: Market-1501 \cite{Zheng_2015_Market1501}, MSMT17 \cite{Wei_2018_MSMT17}.
{\bf Market-1501} consists of 32,668 annotated images of 1,501 identities shot from 6 cameras in total, for which 12,936 images of 751 identities are used for training and 19,732 images of 750 identities are in the test set.
{\bf MSMT17} is the largest Re-ID dataset consisting of 126,441 bounding boxes of 4,101 identities taken by 12 outdoor and 3 indoor cameras.
The training set has 32,621 images of 1,041 identities and the test set has 93,820 images of 3,060 identities.

\textbf{Evaluation protocol.}
In all of the experiments, no ground truth identities were provided for training.
Both Cumulative Matching Characteristics (CMC) \cite{Gray_2007_CMC} Rank1, Rank5, Rank10 accuracies and mean Average Precision (mAP) \cite{Bai_2017_mAP} are used in our experiments. There are no post-processing operations in testing, such as reranking \cite{Zhong_2017_Reranking}.

\subsection{Implementation Details}
\textbf{General training settings}. 
We adopt ResNet-50 \cite{He_2016_ResNet} as the backbone of the feature extractor and initialize the model with the parameters pre-trained on ImageNet \cite{Deng_2009_ImageNet}.
We remove the sub-modules after the "Conv5" and add the global average pooling (gap) followed by the batch normalization and L2 normalization layers, finally it will produce 2048-dimensional features for each image.
An Adam optimizer with a weight decay rate of 0.0005 is used to optimize our networks.
The initial learning rate is 0.00035 with a warmup scheme in the first 10 epochs, the total epoch number is 80 and 50 for Market-1501 and MSMT17 respectively. No learning rate decay is used in the training.

Currently, in the field of unsupervised person Re-ID, two representative papers, CaCL\cite{Lee_2023_CaCL} and DUCL\cite{Liu_2023_DUCL}, adopt the concept of curriculum learning. However, their emphasis differs from the approach proposed in this article. First, the focus is on the field of unsupervised domain adaptation, while the method proposed in this chapter primarily targeting purely unsupervised scenarios. Second, CaCL introduces a camera-driven curriculum learning framework, leveraging camera labels of person images to gradually transfer knowledge from the source to target domains. Specifically, the target domain data is divided into multiple subsets based on the camera labels. Initially, a single subset (images captured by a single camera) is used to train the model. Subsequently, additional subsets are gradually incorporated for training following the course sequence determined by scheduling. The scheduler considers the maximum mean difference (MMD) between each subset and the source domain dataset, so that subsets closer to the source domain enter the course first. DUCL proposes a dual uncertainty-guided course learning framework, aiming to solve two main problems in clustering-based unsupervised learning methods: ignoring the differences in the model's tolerance to various noise levels, which can lead to the model possibly memorizing the error patterns caused by noise, and converging too quickly in the early stages of training, resulting in overfitting. The methods proposed in the first two papers are innovations within the curriculum learning paradigm, while the adaptive intra-class variation contrastive learning algorithm proposed in this paper is inspired by the concept of curriculum learning but does not strictly adhere to the curriculum learning paradigm.

In particular, experiments have shown that MSMT17 and Market1501 cannot use the exact same memory dictionary update strategy. This is because the MSMT17 dataset is more significantly impacted by lighting and weather conditions, and the features other than pedestrians are more complex compared to Market1501. Therefore, the features extracted by the model will be affected by more external features when clustered. When training the MSMT17 dataset, we assign a weight to slow down the sample filtering process.
\begin{equation}
\gamma = \frac{cur\_epoch}{total\_epoch}
\label{msmt_weight}
\end{equation}
where add this weight to \ref{difference}

Our network is trained on 4 NVIDIA GeForce RTX 2080 Ti GPUs using the PyTorch framework. 

\textbf{Training data organization.}
The input image is resized 256 $\times$ 128.
For training images, the input images are resized to 256 $\times$ 128 on all datasets, and the batch size is set to 256. For each mini-batch, we randomly select 16 pseudo identities, and each identity has 16 different images.
Random flipping, cropping, and erasing are applied as data augmentation. 

\begin{table}[t]
\resizebox{\columnwidth}{!}
{
\begin{tabular}{c|c|cc|cc}
\hline
\multirow{2}{*}{Update} & \multirow{2}{*}{Outliers} & \multicolumn{2}{c|}{Market-1501} & \multicolumn{2}{c}{MSMT17} \\ \cline{3-6} 
                        &                           & mAP        & Top-1              & mAP     & Top-1            \\ \hline
CM                      & None                      & 84.9       & 93.7               & 31.7    & 59.8             \\ \hline
Hardest                 & None                      & 85.6       & 94.4               & 19.5    & 40.8             \\ \hline
Linear                  & None                      & 85.1       & 93.8               & 33.3    & 64.0             \\ \hline
Adaptive                & None                      & 86.0       & 93.7               & 32.7    & 62.4             \\ \hline
Adaptive                & All                       & 85.9       & 94.0               & 32.5    & 62.6             \\ \hline
Adaptive                & Adaptive                  & \textbf{86.1}       & \textbf{94.5}               & \textbf{36.0}    & \textbf{65.9}             \\ \hline
\end{tabular}
}
\caption{The effectiveness of each component in our Adaptive Intra-Class Variation Contrastive Learning (AdaInCV). Our method includes the Adaptive Sample Mining and Adaptive Outlier Filter. \textbf{Note}:All results in the ablation experiment were based on the same code framework, with the only modification being the replacement of the memory update module and the outlier module.
}
\label{tab:abla}
\end{table}

\textbf{Clustering settings.}
For the beginning of each epoch, we use DBSCAN \cite{Ester_1996_DBSCAN} for clustering in order to generate pseudo labels. The maximum distance between two samples in DBSCAN is set to 0.5 in Market-1501 and 0.7 in MSMT17.

\subsection{Comparison with State-of-the-art Methods}

We compare our proposed method with state-of-the-art Re-ID methods including the unsupervised domain adaptation methods and the purely unsupervised methods. The comparison results on Market-1501 \cite{Zheng_2015_Market1501} and MSMT17 \cite{Wei_2018_MSMT17} are reported in \ref{tab:sota}.
As illustrated in this table, our proposed method achieves a new state-of-the-art performance of 87.4\%, 38.8\% on the Market-1501 and MSMT17 datasets, respectively. This surpasses all published results for unsupervised person Re-ID tasks on Market-1501 and all methods that do not specifically operate on the camera on MSMT17. Compared with the baseline method HDCRL \cite{Cheng_2022_HDCRL}, our proposed method achieves a higher mAP by \textbf{2.9\%}, and \textbf{18.1\%} on Market-1501 and MSMT17, respectively. Furthermore, when compared to approaches based on contrastive learning, our method demonstrates superior performance on both datasets. Furthermore, our proposed method achieves better performance than all UDA methods that use additionally labeled source datasets on both datasets, even without any identity annotation. Compared with the related state-of-the-art work ISE \cite{Zhang_2022_ISE} which requires generating support samples from actual samples to their neighboring cluster centroids, we can surpass it without using any extra generated samples. With GeM \cite{Radenovic_2018_Gem}, our method still consistently achieves better results. 

Note that, unlike ICE \cite{Chen_2021_ICE} and CAP \cite{Wang_2021_CAP}, we do not utilize the camera information. Under the non-camera setting, our method obtains 36.0\%/65.6\% mAP/Top-1 on MSMT17, largely outperforming ICE \cite{Chen_2021_ICE} (without cameras) and HDCRL \cite{Cheng_2022_HDCRL}.

\subsection{Ablation Study}

\begin{figure}
  \centering{}
  \includegraphics[width=0.36\textheight]{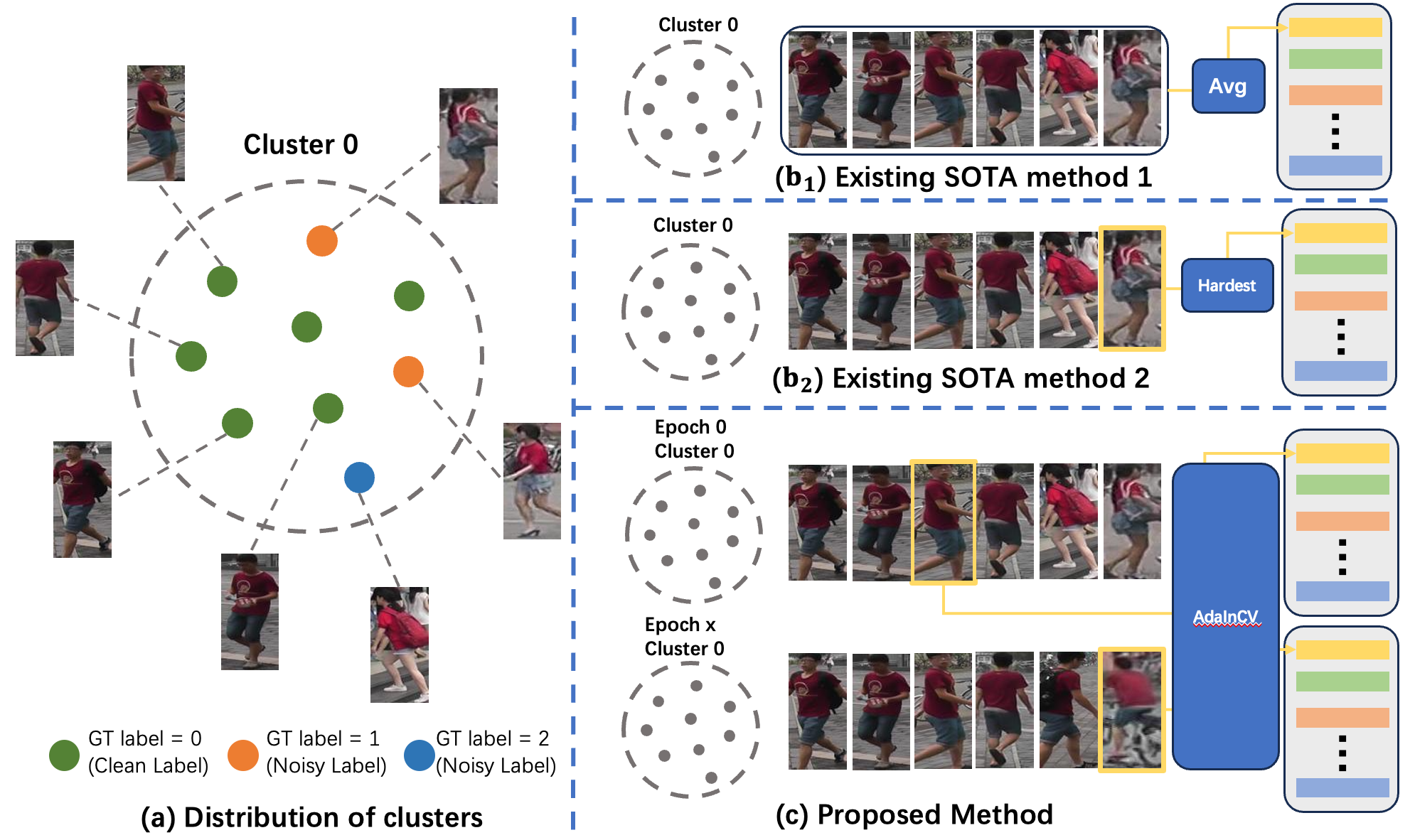}
  \caption{Schematic diagram of Memory Bank update using the method proposed in this paper and the existing SOTA method}
  \vspace{-0.4cm}
  \label{fig:method compare}
\end{figure}

The performance boost of Adaptive Curriculum Contrastive Learning in unsupervised Re-ID mainly comes from the proposed strategies of Adaptive Sample Mining and Adaptive Outliers Filter. We conduct ablation experiments to validate the effectiveness of each component, which is reported in \ref{tab:abla}.

\textbf{Effect of the Adaptive Sample Mining.}
We compared four memory bank update strategies on the Market1501 and MSMT17 datasets. The results show that our method selects different samples from different classes to update the corresponding Memory bank. This is done by calculating the model's learning capabilities for each category. Without considering outliers, our proposed Adaptive Sample Mining method achieved state-of-the-art (SOTA) performance on both datasets, with an increase of 1.1\% and 1.0\% respectively in mean Average Precision (mAP). In particular, CM refers to updating the memory by utilizing all encoding features of the class. Hardest refers to updating the memory by using the features of the hardest instance with the lowest cosine similarity to the query image. Linear refers to using the linear growth strategy in curriculum learning, where samples are selected in an easy-to-difficult manner for memory update. Although both the Linear method and the Adaptive method draw inspiration from curriculum learning implementation, experiments have shown that the samples selected by the Adaptive method are more appropriate than those selected by the Linear method. Only by selecting samples that are compatible with the capabilities of the model can the learning direction of the model be more accurate. Selecting only the most difficult samples will mislead the model in the early stages of learning, and selecting only the simplest samples will not improve the model in the later stages.

\textbf{Effect of the Adaptive Outliers Filter.}

We conducted three main comparative experiments, which involved the following: not adding outlier points, adding all outlier points, and adding outlier points after adaptive filtering. The table above displays the effectiveness of the Adaptive Outliers Filter, which ultimately achieved 86.1\% and 36.0\% on the Market-1501 and MSMT17 datasets, respectively. This shows that although outliers cannot be successfully clustered in the current epoch, selecting appropriate outliers as negative samples to add to the Memory Bank is beneficial for improving model performance. The correct utilization of outliers can enhance our model's generalization and robustness.

\begin{table}[t]\centering
\resizebox{\columnwidth}{!}{
\begin{tabular}{c|ccc|ccc}
\hline
\multirow{2}{*}{Methods} & \multicolumn{3}{c|}{Market1501} & \multicolumn{3}{c}{MSMT17} \\ \cline{2-7} 
                         & epoch     & iters    & mAP     & epoch   & iters   & mAP   \\ \hline
SpCL \cite{Ge_2020_SpCL}           & 50              & 400           & 73.1          & 50            & 800           & 19.1          \\ \hline
ICE \cite{Chen_2021_ICE}           & 40              & 400           & 79.5          & 40            & 400           & 29.8          \\ \hline
CC \cite{Dai_2021_CC}              & \textbf{50}     & \textbf{200}  & \textbf{82.6} & 50            & 400           & 27.6          \\ \hline
ISE \cite{Zhang_2022_ISE}          & 70              & 200           & 84.7          & 50            & 400           & 35.0          \\ \hline
HDCRL \cite{Cheng_2022_HDCRL}      & 120             & 200           & 84.5          & 120           & 400           & 20.7          \\ \hline
Ours                              & 70              & 200           & 87.4          & \textbf{30}   & \textbf{400}  & \textbf{38.8}  \\ \hline
\end{tabular}
}
\caption{
In various unsupervised pedestrian re-identification works that utilize contrastive learning, the epochs between different works and the iters within each epoch were compared.
}
\label{tab:convergence}
\end{table}

We evaluate the impact of our method on the convergence speed of models using the Market-1501 and MSMT17 datasets. From the \ref{tab:convergence}, we can clearly see that our method significantly improves the convergence speed and performance of the model, particularly on the challenging MSMT17 dataset. This demonstrates that our method is capable of learning more effectively in a shorter amount of time, resulting in improved convergence speed and model performance. This increases the possibility of rapid iteration and deployment of this model in the future, in real-life scenarios.

\section{Conclusion}
In this paper, we propose a novel method for unsupervised person Re-ID called adaptive curriculum contrastive learning. The core of this method is an Adaptive Intra-Class Variation Contrastive Learning (\textbf{AdaInCV}) algorithm based on the concept of curriculum learning. This algorithm utilizes the intra-class variation size after clustering to assess the learning capability of the model for each class. This allows for the selection of suitable samples during the model training process. 
Based on this algorithm, we propose two new strategies: Adaptive Sample Mining (\textbf{AdaSaM}) and Adaptive Outlier Filter (\textbf{AdaOF}). AdaSaM evaluates the capabilities of the current model and dynamically selects the most appropriate samples to update the memory dictionary. This allows the model to continue receiving valuable learning signals. AdaOF can dynamically filter out a large number of information-rich outliers as negative samples, thereby enhancing contrastive learning.
Through extensive experiments on two benchmark datasets, our method demonstrates superior performance compared to all existing purely unsupervised and UDA-based Re-ID methods. Furthermore, our method converges faster, which offers significant advantages for future practical applications and deployments.
In fact, this method essentially improves contrastive learning for fine-grained datasets. In the future, we will strive to expand this approach to encompass a broader range of contrastive learning and further explore methods for dynamically acquiring model capabilities.

\bibliographystyle{ACM-Reference-Format}
\bibliography{AdaInCV}

\end{document}